\documentclass[conference]{IEEEtran}
\IEEEoverridecommandlockouts
\usepackage{cite}
\usepackage{amsmath,amssymb,amsfonts}
\usepackage{algorithmic}
\usepackage{graphicx}
\usepackage{textcomp}
\usepackage{xcolor}
\usepackage{float}
\usepackage{color}

\usepackage{soul}

\def\BibTeX{{\rm B\kern-.05em{\sc i\kern-.025em b}\kern-.08em
    T\kern-.1667em\lower.7ex\hbox{E}\kern-.125emX}}
\begin{document}

\title{Building a Massive Corpus for Named Entity Recognition using Free Open Data Sources}

\author{\IEEEauthorblockN{Daniel Specht Menezes}
\IEEEauthorblockA{\textit{Departamento de Inform\'atica} \\
\textit{PUC-Rio}\\
Rio de Janeiro, Brazil \\
danielssmenezes@gmail.com}
\and
\IEEEauthorblockN{Pedro Savarese}
\IEEEauthorblockA{\textit{TTIC} \\
Chicago, USA \\
savarese@ttic.edu}
\and
\IEEEauthorblockN{Ruy L. Milidi\'u}
\IEEEauthorblockA{\textit{Departamento de Inform\'atica} \\
\textit{PUC-Rio}\\
Rio de Janeiro, Brazil \\
milidiu@inf.puc-rio.br}
}

\maketitle

\begin{abstract}

With the recent progress in machine learning, boosted by techniques such as deep learning, many tasks can be successfully solved once a large enough dataset is available for training. Nonetheless, human-annotated datasets are often expensive to produce, especially when labels are fine-grained, as is the case of Named Entity Recognition (NER), a task that operates with labels on a word-level.

In this paper, we propose a method to automatically generate labeled datasets for NER from public data sources by exploiting links and structured data from DBpedia and Wikipedia. Due to the massive size of these data sources, the resulting dataset -- SESAME \footnote{Available at https://sesame-pt.github.io} -- is composed of millions of labeled sentences. We detail the method to generate the dataset, report relevant statistics, and design a baseline using a neural network, showing that our dataset helps building better NER predictors.

\end{abstract}

\section{Introduction}

The vast amounts of data available from public sources such as Wikipedia can be readily used to pre-train machine learning models in an unsupervised fashion -- for example, learning word embeddings \cite{word2vec}. However, large labeled datasets are still often required to successfully train complex models such as deep neural networks, collecting them remain an obstacle for many tasks.

In particular, a fundamental application in Natural Language Processing (NLP) is Named Entity Recognition (NER), which aims to delimit and categorize mentions to entities in text. Currently, deep neural networks present state-of-the-art results for NER, but require large amounts of annotated data for training.

Unfortunately, such datasets are a scarce resource whose construction is costly due to the required human-made, word-level annotations. In this work we propose a method to construct labeled datasets \textbf{without human supervision} for NER, using public data sources structured according to Semantic Web principles, namely, DBpedia and Wikipedia.

Our work can be described as constructing a massive, weakly-supervised dataset (i.e. a silver standard corpora). Using such datasets to train predictors is typically denoted \textit{distant learning} and is a popular approach to training large deep neural networks for tasks where manually-annotated data is scarce. Most similar to our approach are \cite{wiki_ner_joel_1} and \cite{ner_wiki_portugues}, which automatically create datasets from Wikipedia -- a major difference between our method and \cite{ner_wiki_portugues} is that we use an auxiliary NER predictor to capture missing entities, yielding denser annotations.

Using our proposed method, we generate a new, massive dataset for Portuguese NER, called SESAME (\textbf{S}ilv\textbf{e}r-\textbf{S}tandard N\textbf{am}ed \textbf{E}ntity Recognition dataset), and experimentally confirm that it aids the training of complex NER predictors.

The methodology to automatically generate our dataset is presented in Section \ref{sec:method}. Data preprocessing and linking, along with details on the generated dataset, are given in Section \ref{sec:preproc}. Section \ref{sec:baseline} presents a baseline using deep neural networks.

\section{Data sources}
We start by defining what are the required features of the public data sources to generate a NER dataset. As NER involves the delimitation and classification of named entities, we must find textual data where we have knowledge about which entities are being mentioned and their corresponding classes. Throughout this paper, we consider an entity class to be either person, organization, or location.

The first step to build a NER dataset from public sources is to first identify whether a text is \textit{about} an entity, so that it can be ignored or not. To extract information from relevant text, we link the information captured by the DBpedia \cite{dbpedia} database to Wikipedia \cite{wikipedia} -- similar approaches were used in \cite{dbpedia_wikipedia_ner}. The main characteristics of the selected data sources, DBpedia and Wikipedia, and the methodology used for their linkage are described in what follows next.

\subsection{Wikipedia}
Wikipedia is an open, cooperative and multilingual encyclopedia that seeks to register in electronic format knowledge about subjects in diverse domains. The following features make Wikipedia a good data source for the purpose of building a NER dataset.

\begin{itemize}
\item \textbf{High Volume} of textual resources built by humans
\item \textbf{Variety} of domains addressed
\item \textbf{Information boxes:} resources that structure the information of articles homogeneously according to the subject
\item \textbf{Internal links:} links a Wikipedia page to another, based on mentions
\end{itemize}

The last two points are key as they capture human-built knowledge about text is related to the named entities. Their relevance is described in more detail ahead.

\subsubsection{Infobox}
\begin{figure}[htbp]
\centerline{\includegraphics[width=0.4\textwidth]{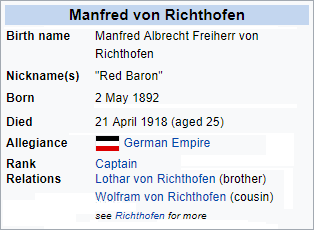}}
\caption{Example of a Wikipedia infobox for a person entity. It consists of a key-value table whose keys depend on the type of the corresponding entity -- for a person entity, common keys include name, birth date, and so on.}
\label{infobox}
\label{fig:infobox}
\end{figure}

Wikipedia infoboxes \cite{infocaixa} are fixed-format tables, whose structure (key-value pairs) are dictated by the article's type (e.g. person, movie, country) -- an example is provided in Figure \ref{fig:infobox}. They present structured information about the subject of the article, and promote structure reuse for articles with the same type. For example, in articles about people, infoboxes contain the date of birth, awards, children, and so on.

Through infoboxes, we have access to relevant human-annotated data: the article's categories, along with terms that identify its subject e.g. name, date of birth. In Figure \ref{infobox}, note that there are two fields that can be used to refer to the entity of the article: "Nickname" and "Birth Name".

Infoboxes can be exploited to discover whether the article's subject is an entity of interest -- that is, a person, organization or location -- along with its relevant details. However, infoboxes often contain inconsistencies that must be manually addressed, such as redundancies e.g. different infoboxes for person and for human. A version of this extraction was done by the DBpedia project, which extracts this structure, and identifies/repairs inconsistencies \cite{dbpediainfobox}.

\subsubsection{Interlinks}
\begin{figure}[htbp]
\centerline{\includegraphics[width=0.45\textwidth]{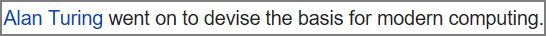}}
\caption{Example of an interlink (blue) in a Wikipedia article. Interlinks point to other articles within Wikipedia, and follow the guideline that only the first mention to the article should contain a link.}
\label{fig:interlink}
\end{figure}

Interlinks are links between different articles in Wikipedia. According to the Wikipedia guidelines, only the first mention to the article must be linked. Figure \ref{fig:interlink} shows a link (in blue) to the article page of Alan Turing: following mentions to Alan Turing in the same article must not be links.

While infoboxes provide a way to discover relevant information about a Wikipedia article, analyzing an article's interlinks provide us access to referenced entities which are not the page's main subject. Hence, we can parse every article on Wikipedia while searching for interlinks that point to an entity article, greatly expanding the amount of textual data to be added in the dataset.

\subsection{DBpedia}
DBpedia extracts and structures information from Wikipedia into a database that is modeled based on semantic Web principles \cite{semantic_web}, applying the Resource Description Framework (RDF).
Wikipedia's structure was extracted and modelled as an ontology \cite{dbpediaontology}, which was only possible due to infoboxes.

The DBpedia ontology focused on the English language and the extracted relationships were projected for the other languages. In short, the ontology was extracted and preprocessed from Wikipedia in English and propagated to other languages using interlinguistic links. Articles whose ontology is only available in one language are ignored.

An advantage of DBpedia is that manual preprocessing was carried out by project members in order to find all the relevant connections, redundancies, and synonyms -- quality improvements that, in general, require meticulous human intervention. In short, DBpedia allows us to extract a set of entities where along with its class, the terms used to refer to it, and its corresponding Wikipedia article.

\section{Building a database}
\label{sec:method}

The next step consists of building a structured database with the relevant data from both Wikipedia and DBpedia.

\subsection{DBpedia data extraction}
Data from DBpedia was collected using a public service access \cite{dbpedia_endpoint}. We searched over the following entity classes: people, organizations, and locations, and extracted the following information about each entity:

\begin{itemize}
\item The entity's class (person, organization, location)
\item The ID of the page (Wiki ID)
\item The title of the page
\item The names of the entity. In this case the ontology varies according to the class, for example, place-type entities do not have the "surname" property
\end{itemize}

\subsection{Wikipedia data extraction}
We extracted data from the same version of Wikipedia that was used for DBpedia, October 2016, which is available as \textit{dumps} in XML format. We extracted the following information about the articles:
\begin{itemize}
\item Article title
\item Article ID (a unique identifier)
\item Text of the article (in wikitext format)
\end{itemize}

\subsection{Database modelling}
\begin{figure}[htbp]
\centerline{\includegraphics[width=0.45\textwidth]{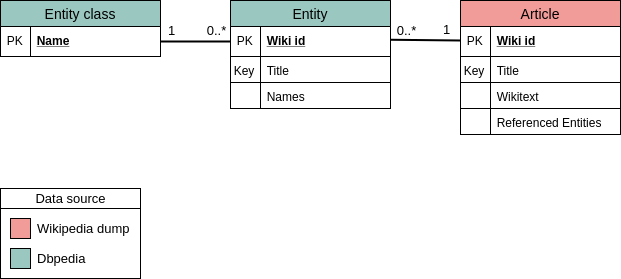}}
\caption{Diagram of the database representing the links between entities and articles.}
\label{db_diagram}
\end{figure}

Figure \ref{db_diagram} shows the structure of the database as a
entity-relation diagram. Entities and articles were linked when either one of two linked articles correspond to the entity, or the article itself is about a known entity.

\section{Preprocessing}
\label{sec:preproc}

\subsection{Wikitext preprocessing}

We are only interested in the plain text of each Wikipedia article, but its Wikitext (language used to define the article page) might contain elements such as lists, tables, and images. We remove the following elements from each article's Wikitext:
\begin{itemize}
\item Lists, (e.g. unbulled list, flatlist, bulleted list)
\item Tables (e.g. infobox, table, categorytree)
\item Files (e.g. media, archive, audio, video)
\item Domain specific (e.g. chemistry, math)
\item Excerpts with irregular indentation (e.g. outdent)
\end{itemize}

\subsection{Section Filtering}
Wikipedia guidelines include sets of suggested sections, such as early life (for person entities), references, further reading, and so on. Some of the sections have the purpose of listing related resources, not corresponding to a well structured text and, therefore, can be removed with the intent to reduce noise. In particular, we remove the following sections from each article: ``references'', ``see also'', ``bibliography'', and ``external links''.

After removing noisy elements, the Wikitext of each article is converted to raw text. This is achieved through the tool MWparser \cite{mwparserfromhell}.

\subsection{Identifying entity mentions in text}

The next step consists of detecting mentions to entities in the raw text. To do this, we tag character segments that exactly match one of the known names of an entity. For instance, we can tag two different entities in the following text:
\begin{align*}
\underbrace{\text{John Smith}}_{\text{PER}}
\text{ travelled to }
\underbrace{\text{Rio de Janeiro}}_{\text{LOC}}
\text{. Visited Copacabana.}
\end{align*}
Note that the word ``Copacabana'' can also correspond to a ``Location'' entity. However, some entity mentions in raw text might not be identified in case they are not present in DBpedia.

\subsection{Searching for other entities}
\label{sec:other_entities}

To circumvent mentioned entities which are not present in DBpedia, we use an auxiliary NER system to detect such mentions. More specifically, we use the Polyglot \cite{polyglot} system, a model trained on top of a dataset generated from Wikipedia.

Each mention's tag also specifies whether the mention was detected using DBpedia or by Polyglot. The following convention was adopted for the tags:
\begin{itemize}
\item Annotated (Anot) - Matched exactly with one of the the entity's names in DBpedia
\item Predicted (Pred) - Extracted by Polyglot
\end{itemize}
 
 Therefore, in our previous example, we have:
\begin{align*}
\underbrace{\text{John Smith}}_{\text{Anot-PER}}
\text{ travelled to }
\underbrace{\text{Rio de Janeiro}}_{\text{Anot-LOC}}
\text{. Visited }
\underbrace{\text{Copacabana}}_{\text{Pred-LOC}}
\text{.}
\end{align*}
A predicted entity will be discarded entirely if it conflicts with an annotated one, since we aim to maximize the entities tagged using human-constructed resources as knowledge base.

\subsection{Tokenization of words and sentences}
The supervised learning models explored in this paper require inputs split into words and sentences. This process, called tokenization, was carried with the NLTK toolkit \cite{nltk}, in particular the "Punkt" tokenization tool, which implements a multilingual, unsupervised algorithm \cite{punkt_algorithm}.

First, we tokenize only the words corresponding to mentions of an entity. In order to explicitly mark the boundaries of each entity, we use the BIO format, where we add the suffix ``B'' (begin) to the first token of a mention and ``I'' (inside) to the tokens following it. This gives us:
\\\\
$
\underbrace{\text{John}}_{\text{B-PER}}
\underbrace{\text{Smith}}_{\text{I-PER}}
\text{travelled to}
\underbrace{\text{Rio}}_{\text{B-LOC}}
\underbrace{\text{de}}_{\text{I-LOC}}
\underbrace{\text{Janeiro}}_{\text{I-LOC}}
\text{. Visited }
\underbrace{\text{Copacabana}}_{\text{B-LOC}}
$
\\\\
Second, we tokenize the remaining text, as illustrated by the following example: $w_{i}$ denotes a word token, while $s_{i}$ corresponds to a sentence token.
\begin{align*}
\underbrace{\underbrace{\underbrace{\text{John}}_{\text{B-PER}}}_{\displaystyle w_{0}}
\underbrace{\underbrace{\text{Smith}}_{\text{I-PER}}}_{\displaystyle w_{1}}
\underbrace{\color{white}\underbrace{\color{black}\text{travelled}}_{\text{O}}}_{\displaystyle w_{2}}
\underbrace{\color{white}\underbrace{\color{black}\text{to}}_{\text{O}}}_{\displaystyle w_{3}}
\underbrace{\underbrace{\text{Rio}}_{\text{B-LOC}}}_{\displaystyle w_{4}}
\underbrace{\underbrace{\text{de}}_{\text{I-LOC}}}_{\displaystyle w_{5}}
\underbrace{\underbrace{\text{Janeiro}}_{\text{I-LOC}}}_{\displaystyle w_{6}}
\underbrace{\color{white}\underbrace{\color{black}\text{.}}_{\text{O}}}_{\displaystyle w_{7}}}_{\displaystyle s_{0}}
\end{align*}
However, conflicts might occur between known entity tokens and the delimitation of words and sentences. More specifically, tokens corresponding to an entity must consist only of entire words (instead of only a subset of the characters of a word), and must be contained in a single sentence. In particular, we are concerned with the following cases:

(1) Entities which are not contained in a single sentence:
\begin{align*}
\underbrace{\underbrace{w_{0}}_{\text{O}}
\underbrace{w_{1}}_{\text{B-PER}}}_{\displaystyle s_{0}}
\underbrace{\underbrace{w_{2}}_{\text{I-PER}}
\underbrace{w_{3}}_{\text{O}}}_{\displaystyle s_{1}}
\end{align*}
In this case, $w_{1}$ and $w_{2}$ compose a mention of the entity which lies both in sentence $s_{0}$ and $s_{1}$. Under these circumstances, we concatenate all sentences that contain the entity, yielding, for the previous example:
\begin{align*}
\underbrace{\underbrace{w_{0}}_{\text{O}}
\underbrace{w_{1}}_{\text{B-PER}}
\underbrace{w_{2}}_{\text{I-PER}}
\underbrace{w_{3}}_{\text{O}}}_{\displaystyle s_{0}}
\end{align*}
(2) Entities which consist of only subsets (some characters) of a word, for example:
\begin{align*}
\lefteqn{\overbrace{\phantom{\ c_{0}\ c_{1}\ c_{2}}}^{\displaystyle w_{0}}}\ c_{0}\ c_{1}
\underbrace{c_{2} \
\lefteqn{\overbrace{\phantom{\ c_{3}\ c_{4}\ c_{5}}}^{\displaystyle w_{1}}}\ c_{3}}_{\text{B-PER}}\ c_{4}\ c_{5}
\end{align*}
In this case, we remove the conflicting characters from their corresponding word tokens, resulting in:
\begin{align*}
\overbrace{c_{0}\ c_{1}}^{\displaystyle w_{0}}
\underbrace{c_{2}\ c_{3}}_{\text{B-PER}}
\overbrace{c_{4}\ c_{5}}^{\displaystyle w_{1}}
\end{align*}

\subsection{Dataset structure}
The dataset is characterized by lines corresponding to words extracted from the preprocessing steps described previously, following the BIO annotations methodology.

Each word is accompanied with a corresponding tag, with the suffix PER, ORG or LOC for person, organization, and location entities, respectively. Moreover, word tags have the prefix "B" (begin) if the word is the first of an entity mention, "I" (inside) for all other words that compose the entity, and "O" (outside) if the word is not part of any entity. Blank lines are used to mark the end of an sentence. An example is given in Table \ref{exemplo_bio}.

\begin{table}[H]
\caption{Example of a sentence ("John Smith went to Rio de Janeiro") following the structure of the proposed dataset, where each line consists of a token and a corresponding tag in BIO form.}
\label{exemplo_bio}
\centering
\renewcommand{\tablename}{Exemplo}
\begin{tabular}{rl}
John    & B-PER \\
Smith  & I-PER \\
went     & O \\
to      & O \\
Rio     & B-LOC \\
de      & I-LOC \\
Janeiro & I-LOC\\
.       & O \\
\end{tabular}
\end{table}

\subsection{Semantic Model}

Since our approach consists of matching raw text to a list of entity names, it does not account for context in which the entity was mentioned. For example, while under a specific context a country entity can exert the role of an organization, our method will tag it as a location regardless of the context. Therefore, our approach delimits an entity mention as a semantic object that does not vary in according to the context of the sentence.

\subsection{SESAME}
By following the above methodology on the Portuguese Wikipedia and DBpedia, we create a massive silver standard dataset for NER. We call this dataset SESAME (\textbf{S}ilv\textbf{e}r-\textbf{S}tandard N\textbf{am}ed \textbf{E}ntity Recognition dataset). We then proceed to study relevant statistics of SESAME, with the goal of:

\begin{enumerate}
\item Acknowledging inconsistencies in the corpus, e.g. sentence sizes
\item Raising information relevant to the calibration and evaluation of model performance e.g. proportion of each entity type and of each annotation source (DBpedia or auxiliary NER system)
\end{enumerate}

We only consider sentences that have annotated entities. After all, sentences with only parser extraction entities do not take advantage of the human discernment invested in the structuring of the data of DBpedia.

\subsubsection{Sentences}
\begin{figure}[htbp]
\centerline{\includegraphics[width=0.45\textwidth]{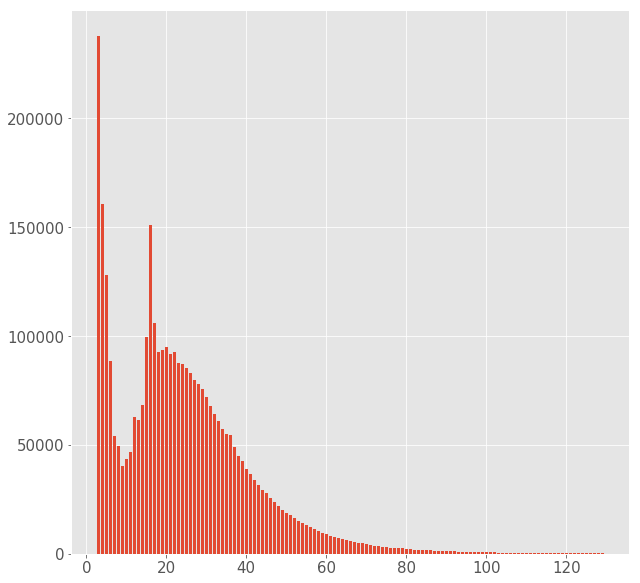}}
\caption{Number of occurrences for different sentence lengths (number of tokens) in the generated corpus, with a mean of approximately 24 words.}
\label{sentence_size}
\end{figure}

SESAME consists of \textbf{3,650,909} sentences, with lengths (in terms of number of tokens) following the distribution shown in Figure \ref{sentence_size}. A breakdown of relevant statistics, such as the mean and standard deviation of sentences' lengths, is given in Table \ref{table_tamanho_sentencas}.

\begin{table}[htbp]
\caption{Statistics on the length of sentences, measured in number of tokens, across SESAME.}
\begin{center}
\begin{tabular}{ll}
\hline
Metric       &         \\ \hline
Total         & 3,650,909 \\
$\mu$ Mean         & 24.04   \\
$\sigma$ Standard deviation  & 20.41   \\
Min        & 1       \\
Max        & 8437    \\
$Q_{1}$  Percentile 25\%          & 11      \\
$Q_{2}$ Percentile 50\%          & 21      \\
$Q_{3}$ Percentile 75\%          & 33      \\ \hline
\end{tabular}
\label{table_tamanho_sentencas}
\end{center}
\end{table}

\subsubsection{Tokens}
SESAME consists of \textbf{87,769,158} tokens in total. The count and proportion of each entity tag (not a named entity, organization, person, location) is given in \ref{table_statistics_tags_all}.

\begin{table}[htbp]
\caption{Frequency of each tag across SESAME.}
\begin{center}
\begin{tabular}{lll}
\hline
Class        & Total & \%  \\ \hline
Not NE       &  81,357,679 & 92.69\% \\
Organization & 1,053,298 & 1.20\%  \\
Person       & 1,878,838 & 2.14\%  \\
Location     & 3,479,343 & 3.96\%  \\ \hline
\end{tabular}
\label{table_statistics_tags_all}
\end{center}
\end{table}

Not surprisingly, the vast majority of words are not related to an entity mention at all. The statistics among words that are part of an entity mention are given in Table \ref{table_statistics_tags_entities}, where over half of the entity mentions are of the type location.

\begin{table}[htbp]
\begin{center}
\caption{Frequency of each entity tag (all except for "O", "outside") across SESAME.}
\label{table_statistics_tags_entities}
\begin{tabular}{lll}
\hline
Class        & Total & \%  \\ \hline
Organization & 1,053,298 & 16.42\%  \\
Person       & 1,878,838 & 29.30\%  \\
Location     & 3,479,343 & 54.26\%  \\ \hline
\end{tabular}
\end{center}
\end{table}



Table \ref{corpus_comparison} shows a size comparison between SESAME and popular datasets for Portuguese NER.

\begin{table}[htbp]
\begin{center}
\caption{Comparison between NER datasets for Portuguese.}
\label{corpus_comparison}
\begin{tabular}{lll}
\hline
Corpus          & Sentences & Tokens \\ \hline
SESAME          &   3,650,909   &   87,769,158 \\ 
Paramopama      &   12,500      &   310,000 \\
First  HAREM &   5,000       &   80,000 \\
Second  HAREM  &   3,500       &   100,000 \\
WikiNER         &   125,821     &   2,830,000 \\
\hline
\end{tabular}
\end{center}
\end{table}

\subsubsection{Entity origin}
Table \ref{table_origin} presents the proportion of matched and detected mentions for each entity type -- recall that tagged mentions have either been matched to DBpedia (hence have been manually annotated) or have been detected by the auxiliary NER system Polyglot.

As we can see, the auxiliary predictor increased the number of tags by $33\%$ relatively, significantly increasing the number of mentions of type organization and person -- which happen to be the least frequent tags.

\begin{table}[htbp]
\begin{center}
\caption{ Frequency of tokens originated from annotation and from detection by the auxiliary NER predictor.}
\label{table_origin}
\begin{tabular}{lll}
\hline
Class        & Annotated & Detected  \\ \hline
All Classes  &   75.60\%     &   24.39\% \\
Organization &   67.69\%     &   32.30\% \\
Person       &   65.76\%     &   34.23\% \\
Location     &   83.31\%     &   16.68\% \\ \hline
\end{tabular}
\end{center}
\end{table}

\section{Baseline}
\label{sec:baseline}

To construct a strong baseline for NER on the generated SESAME dataset and validate the quality of datasets generated following our method, we use a deep neural network that proved to be successful in many NLP tasks. Furthermore, we check whether adding the generated corpus to its training dataset provides performance boosts in NER benchmarks.

\subsection{Datasets}
In order to have a fair evaluation of our model, we use human-annotated datasets as validation and test sets.

We use the first HAREM and miniHAREM corpus, produced by the Linguateca project \cite{linguateca}, as gold standard for model evaluation. We split the dataset in the following manner:
\begin{itemize}
\item Validation: 20\% of the first HAREM
\item Test: 80\% of the first HAREM, plus the mini HAREM corpus
\end{itemize}

Another alternative is to use the Paramopama corpus which is larger than the HAREM and miniHAREM datasets. However, it was built using an automatic refinement process over the WikiNER corpus, hence being a silver standard dataset. We opted for the smaller HAREM datasets as they have been manually annotated, rendering the evaluation fair.

The HAREM corpus follows a different format than the one of SESAME: it uses a markup structure, without a proper tokenization of sentences and words. To circumvent this, we convert it to BIO format by applying the same tokenization process used for generating our dataset.

\subsection{Evaluation}
The standard evaluation metric for NER is the  $F_1$ score: 
\begin{equation*}
    F_1 = 2 \cdot \frac{P \cdot R}{P + R}
\end{equation*}
where $P$ stands for precision and R for recall. \textit{Precision} is the percentage of entity predictions which are correct, while \textit{Recall} is the percentage of entities in the corpus that are correctly predicted by the model.

Instead of the standard $F_1$ score, we follow the evaluation proposed in \cite{harem_1}, which consists of a modified \textit{First HAREM} $F_1$ score used to compare different models. Our choice is based on its wide adoption in the Portuguese NER literature \cite{character_level_ner}, \cite{ETLCMT}, \cite{sistema_cortex}.

In particular, for the \textit{First HAREM} $F_1$ score: (1) as the corpus defines multiple tags for the same segments of the text, the evaluation also accepts multiple correct answers; (2) partial matches are considered and positively impact the score.

In this work, the configuration of the First HAREM evaluation procedure only considers the classes ``person'', ``location'' and ``organization''. Also, the HAREM corpus has the concept of ``subtype'' e.g. an entity of the type  ``person'' can have the subtype ``member''. We only perform evaluation considering the main class of the entity.


\subsection{Baseline results}

We performed extensive search over neural network architectures along with grid search over hyperparameter values. The model that yielded the best results consists of: (1) a word-level input layer, which computes pre-trained word embeddings \cite{glove} along with morphological features extracted by a character-level convolutional layer \cite{character_level}, (2) a bidirectional LSTM \cite{bi_lstm}, (3) two fully-connected layers, and (4) a conditional random field (CRF). Table \ref{hyperparameters} contains the optimal found hyperparameters for the network.

\begin{table}[htbp]
\begin{center}
\caption{Hyperparameters for the deep network that yielded optimal results for NER on SESAME, found through grid search.}
\label{hyperparameters}

\begin{tabular}{l|l}
\hline
\multicolumn{2}{l}{Word embeddings}                       \\ \hline
Embeddings (pre-trained)              & GloVe                               \\
Size                & 100                                 \\ \hline
\multicolumn{2}{l}{Character-level convolution}              \\ \hline
Window size         & 3                                   \\
Filters             & 30                                  \\
Dropout (output)             & 0.5 \\ \hline
\multicolumn{2}{l}{biLSTM}                                  \\ \hline
Hidden Units               & 150                                 \\
Variational Dropout & 0.25                                 \\
Dropout (input)             & 0.5              \\ \hline
\multicolumn{2}{l}{Fully-connected Layers}                          \\ \hline
Depth             & 2                                   \\
Activation          & tanh                                \\
Hidden units                & 100                                 \\ \hline
\multicolumn{2}{l}{Optimization}                 \\ \hline
Optimizer           & SGD                                 \\
Momentum            & 0.9                                 \\
Gradient clipping   & 5.0                                 \\
Mini-batch size     & 10                                  \\
Learning rate       & 0.005                              
\end{tabular}
\end{center}
\end{table}

Additionally, the baseline was developed on a balanced re-sample of SESAME with a total of 1,216,976 sentences. The model also receives additional categorical features for each word, signalizing whether it: (1) starts with a capital letter, (2) has capitalized letters only, (3) has lowercase letters only, (4) contains digits, (5) has mostly digits ($>$ 50\%) and (6) has digits only.

With the goal of evaluating whether SESAME can be advantageous for training NER classifiers, we compare the performance of the neural network trained with and without it. More specifically, we train neural networks on the HAREM2 \cite{harem_2} dataset, on SESAME, and on the union of the two -- Table \ref{results} shows the test performance on the first HAREM corpus. As we can see, while SESAME alone is not sufficient to replace a human-annotated corpus (the $F_1$ score of the network trained on the SESAME is lower than the one trained on the HAREM2 corpus), it yields a boost of $1.5$ in the $F_1$ score when used together with the HAREM2 dataset.

\begin{table}[htbp]
\begin{center}
\caption{Baseline results: using SESAME along with a human-annotated corpus boosts $F_1$ performance.}
\label{results}
\begin{tabular}{c|ccc}
\hline
Training Data   & $F_1$             & Precision                 & Recall                    \\ \hline
SESAME   & \multicolumn{1}{c}{67.49} & \multicolumn{1}{c}{77.53} & \multicolumn{1}{c}{59.76} \\
HAREM2  &   \multicolumn{1}{c}{72.72} & \multicolumn{1}{c}{75.28} & \multicolumn{1}{c}{70.32} \\
SESAME + HAREM2  & \multicolumn{1}{c}{\textbf{74.22}} & \multicolumn{1}{c}{\textbf{77.58}} & \multicolumn{1}{c}{\textbf{71.14}} \\ \hline
    
\end{tabular}
\end{center}
\end{table}

\section{Conclusion}

Complex models such as deep neural networks have pushed progress in a wide range of machine learning applications, and enabled challenging tasks to be successfully solved. However, large amounts of human-annotated data are required to train such models in the supervised learning framework, and remain the bottleneck in important applications such as Named Entity Recognition (NER). We presented a method to generate a massively-sized labeled dataset for NER in an automatic fashion, without human labor involved in labeling -- we do this by exploiting structured data in Wikipedia and DBpedia to detect mentions to named entities in articles. 

Following the proposed method, we generate SESAME, a dataset for Portuguese NER. Although not a gold standard dataset, it allows for training of data-hungry predictors in a weakly-supervised fashion, alleviating the need for manually-annotated data. We show experimentally that SESAME can be used to train competitive NER predictors, or improve the performance of NER models when used alongside gold-standard data. We hope to increase interest in the study of automatic generation of silver-standard datasets, aimed at distant learning of complex models. Although SESAME is a dataset for the Portuguese language, the underlying method can be applied to virtually any language that is covered by Wikipedia.


\bibliographystyle{IEEEtran}
\bibliography{refs}
\end{document}